\documentclass{sig-alternate}

\usepackage[latin1]{inputenc}
\usepackage{amsmath}
\usepackage{amsfonts}
\usepackage{amssymb}
\usepackage{mathrsfs}
\usepackage{xspace}
\usepackage{graphicx} 
\usepackage[tight]{subfigure}
\usepackage{color}
\usepackage{multirow}

\usepackage{amsfonts}
\usepackage[sort&compress,square,comma,numbers]{natbib}
\usepackage{algorithm}
\usepackage{algorithmic}
\newcommand{\algo}{{\textsc{GP-Select}}\xspace}
\newcommand{\acron}{{\textsc{AVID}}\xspace}
\newcommand{\acronex}{{\textsc{AVID} - Adaptive Valuable Item Discovery}\xspace}

\usepackage{url}
\usepackage{hyperref}

\usepackage{tikz}
\usetikzlibrary{matrix}
\usepackage{PTSansNarrow}
\usepackage[T1]{fontenc}

\usetikzlibrary{matrix}

\newcommand{\groundSet}{{\mathbf{V}}}
\newcommand{\cand}{{v}}
\newcommand{\subSet}{{\mathit{S}}}
\newcommand{\mFn}{{F}}

\newcommand{\ivFn}{{f}}
\newcommand{\smFn}{{D}}

\newcommand{\gpucb}{{\textsf{GP-UCB}}\xspace}

\newcommand{\vobs}{\mathbf{y}}
\newcommand{\noise}{\epsilon}
\newcommand{\noisevar}{{\sigma_n}}

\newcommand{\eqnref}[1]{(\ref{#1})\xspace}

\newcommand{\kernel}{{\kappa}}
\newcommand{\Kernel}{{\mathbf{K}}}
\newcommand{\whenincapp}[1]{\ifthenelse{\boolean{incapp}}{#1}{}}

\newcommand{\ucb}{{\textsc{GP-UCB}}\xspace}

\newcommand{\commentout}[1]{}

\newtheorem{theorem}{Theorem}
\newtheorem{lemma}{Lemma}

\DeclareMathOperator*{\argmax}{argmax}

\newcommand{\vect}[1]{\boldsymbol{#1}} 
\newcommand{\mat}[1]{\boldsymbol{#1}} 

\newcommand{\dummystring}{QWERTYU}
\newcommand{\vci}[3][\dummystr]{\ifthenelse{\equal{#1}{\dummystring}}{\vect{#2}_{#3}}{\vect{#2}_{#3}^{(#1)}}}
\newcommand{\mx}[3][\dummystr]{\ifthenelse{\equal{#1}{\dummystring}}{\mat{#2}_{#3}}{\mat{#2}_{#3}^{(#1)}}}

\newcommand{\Cov}{\mathrm{Cov}}

\renewcommand{\eqref}[1]{Eq.~\ref{eq:#1}}

\begin{document}


\title{Discovering Valuable Items from Massive Data}

\numberofauthors{5}

\author{
\alignauthor
Hastagiri P Vanchinathan\\
	\affaddr{ETH Zurich}\\
	\email{hastagiri@inf.ethz.ch}\\ 	
\alignauthor
Andreas Marfurt\\
	\affaddr{ETH Zurich}\\
	\email{amarfurt@ethz.ch} \\	
\alignauthor
Charles-Antoine Robelin\\
	\affaddr{Amadeus IT Group SA}\\
	\email{carobelin@amadeus.com}\\ 
\and		
\alignauthor
Donald Kossmann\\
	\affaddr{ETH Zurich}\\
	\email{donaldk@ethz.ch}\\ 	
\alignauthor
Andreas Krause\\
	\affaddr{ETH Zurich}\\
	\email{krausea@ethz.ch}\\ 
}
\maketitle
\begin{abstract}
  Suppose there is a large collection of items, each with an associated cost and an inherent utility that is revealed only once we commit to selecting it. Given a budget on the cumulative cost of the selected items, how can we pick a subset of maximal value? This task generalizes several important problems such as multi-arm bandits, active search and the knapsack problem. We present an algorithm, \algo, which utilizes prior knowledge about similarity between items, expressed as a kernel function. \algo uses Gaussian process prediction to balance exploration (estimating the unknown value of items) and exploitation (selecting items of high value). We extend \algo to be able to discover sets that simultaneously have high utility and are diverse. Our preference for diversity can be specified as an arbitrary monotone submodular function that quantifies the diminishing returns obtained when selecting similar items. Furthermore, we exploit the structure of the model updates to achieve an order of magnitude (up to 40X) speedup in our experiments without resorting to approximations. We provide strong guarantees on the performance of \algo and apply it to three real-world case studies of industrial relevance: (1) Refreshing a repository of prices in a Global Distribution System for the travel industry, (2) Identifying diverse, binding-affine peptides in a vaccine design task and (3) Maximizing clicks in a web-scale recommender system by recommending items to users.
\end{abstract}



\section{INTRODUCTION}
\label{sec:intro}
Consider a large collection of items, each having an inherent value and an associated cost. We seek to select a subset of maximal value, subject to a constraint on the cumulative cost of the selected items.
If we know the items' values and costs, this is just the classical knapsack problem - which is NP-hard, but can be near-optimally solved, e.g.,  using dynamic programming. But what if we do not know the values? Concretely, we consider the setting where we can choose an item, observe a noisy estimate of its value, then choose and evaluate a second item and so on, until our budget is exhausted. It is clear that in order to achieve non-trivial performance, we must be able to make predictions about the value of non-selected items given observations made so far.  Hence, we will assume that we are given some information about the similarity of items (e.g., via features), whereby similar items are expected to yield similar value.
As a motivating application, consider experimental design, where we may need to explore a design space, and wish to identify a set of near-optimal designs, evaluating one design at a time. In the early stages of medical drug development, for example, candidate compounds are subject to various tests and a fixed number of them are selected to the next stage to perform animal/human testing. Even the initial tests are expensive and the goal is to reduce the number of compounds on which these tests are conducted while still selecting a good set of compounds to promote to the next level. 
Another application is recommender systems, where for a given customer, we may seek to iteratively recommend items to read/watch, aiming to maximize the cumulative relevance of the entire set. Alternatively, we might want to pick users from our user base or a social network to promote a given item. In this setting, how should we select items to maximize total utility? 

We will call this general class of problems {\em \acronex}. To solve \acron, we need to address an exploration--exploitation dilemma, where we must select items that maximize utility (exploit) while simultaneously estimating the utility function (explore). 
We 
address these challenges by using ideas from Gaussian Process optimization and multi-armed bandits to provide a principled approach to \acron with strong theoretical guarantees. 
Specifically, we introduce a novel algorithm, \algo, for discovering high value items in a very general setting. \algo can be used whenever the similarity between items can be captured by a positive definite kernel function, and the utility function has low norm in the Reproducing Kernel Hilbert Space (RKHS) associated with the kernel. The algorithm models the utility function as a sample from a Gaussian process distribution, and uses its predictive uncertainty to navigate the exploration--exploitation tradeoff via an upper confidence based sampling approach that takes item costs into account.

We also consider a natural extension of \acron, where the goal is to obtain a \emph{diverse} set of items. This is an important requirement in many experimental design problems where, for example for reasons of robustness, we seek to identify a collection of diverse, yet high quality designs. In our drug design example, very similar compounds might cause similar side effects in the later stages of testing. Hence, we might require a certain diversity in the selected subset while still trying to maximize total value. In this work, we address the setting where our preference for diversity is quantified by a submodular function, modeling diminishing returns incurred when picking many similar items.  We prove that \algo 
provides an effective tradeoff of value and diversity, establishing bounds on its regret against an omniscient algorithm with access to the unknown objective. Our results substantially expand the class of problems that can be solved with upper confidence based sampling methods -- desirable for their simplicity -- in a principled manner.  


We evaluate \algo in three real-world case studies. We first demonstrate how \algo can be used to maintain an accurate repository of ticket prices in a Global Distribution System that serves a large number of airlines and travel agencies. Here the challenge is to selectively recompute ticket prices that likely have changed, under a budget on the number of computations allowed.  Secondly, we demonstrate how \algo is able to determine a diverse set of candidate designs in a vaccine design application exhibiting high binding affinity to their target receptors. In these experiments, we also study the effect of inducing diversity, and non-uniform selection cost. 
Finally, we present results on a web-scale recommender systems dataset provided by Yahoo!~where the task is to adaptively select user-item pairs that maximize interaction (clicks, likes, shares, etc.). 

Our experiments highlight the efficacy of \algo and its applicability to a variety of problems relevant to practitioners. In particular, with our suggested application of lazy variance updates, we are able to speed up the execution by up to almost 40 times, making it usable on web-scale datasets.

\section{AVID: PRELIMINARIES}
\label{sec:approach}
We are given a set $\groundSet = \{ 1,\dots,n\}$ of $n$ objects. 
There is a utility function $\ivFn:\groundSet \rightarrow {\mathbb{R}}_{\geq 0}$ that assigns a non-negative value to every item in the set. Similarly, there is a function $c:\groundSet \rightarrow {\mathbb{R}}_{>0}$, assigning a positive cost $c_{\cand}=c(\cand)\in [c_{min}, c_{max}]$ to each item $\cand$. Given a subset $\subSet \subseteq \groundSet$, its value $\mFn(S) = 
\sum_{\cand \in \subSet} \ivFn(\cand)$ is the sum of the values of the selected items, and its cost $C(S)$ the cumulative costs of the items. Given a budget $B>0$, our goal is to select 
\begin{equation}\subSet_B^*=\argmax_{C(\subSet)\leq B} F(\subSet),\label{eqn:knapsack}\end{equation}
i.e., a subset of maximum value, with cost bounded by $B$.

If we knew  the utility function $f$, then Problem~\eqnref{eqn:knapsack} is the classical knapsack problem. While NP-hard, for any $\varepsilon$, an $\varepsilon$-optimal solution can be found via dynamic programming.

But what if we do not know $f$? In this case, we consider  
choosing a subset $\subSet$ in a sequential manner.
We pick one item at a time, after which the value of the selected item is revealed (possibly perturbed by noise), and can be taken into account when selecting further items. We term this sequential problem {\em \acronex}.

Equivalent to maximizing the cumulative value $\mFn(S)$, we aim to minimize the {\em regret}, i.e., the loss in cumulative value compared to an omniscient optimal algorithm that knows $f$.  
Formally, the regret of a subset $\subSet_B$ of cost $B$ is defined as: $R_B = \mFn (\subSet_B^*) - \mFn (\subSet_B)$. We seek an algorithm whose regret grows slowly (sublinearly) with the budget $B$, so that the average regret $R_B/B$ goes to $0$. 


%

\paragraph{Diversity} In several important applications, we not only seek items of high value, but also to optimize the diversity of the selected set. 
One way to achieve this goal is to add to our objective another term that prefers diverse sets.
Concretely, 
we extend the scope of \acron by considering objective functions of the form: 
\begin{equation}
\mFn(\subSet)= (1-\lambda)\displaystyle\sum_{\cand \in \subSet} \ivFn (\cand) + \lambda \smFn(\subSet).
\label{eqn:DivObjective}
\end{equation}
Hereby, 
$\smFn(S)$ is a {\em known} measure of the diversity of the selected subset $S$. Many such diversity-encouraging objectives have been considered in the literature (c.f., \citep{Streeter09online,Lin11,Yue11,Kulesza12}). We will present an algorithm that is guaranteed to choose near-optimal sets whenever the function $\smFn$ satisfies {\em submodularity}. Submodularity is a natural notion of diminishing returns, capturing the idea that adding an item helps less if more similar items were already picked \citep{Choquet54}. We discuss examples in Section~\ref{sec:diversity}.
$\lambda \in [0,1]$ is a tradeoff parameter balancing the relative importance of value and diversity of the selected set. In the case where $f$ is known, maximizing $\smFn$ requires maximizing a submodular function. This task is NP-hard, but can be solved  near-optimally using a greedy algorithm \citep{Nemhauser78}.  In this paper, we address the novel setting where $D$ is any known submodular function but $f$ is {\em unknown}, and needs to be estimated.

\paragraph{Regularity Assumptions}
\label{subsec:regularity}
In the general case, where $\ivFn$ can be any function, it is hopeless to compete against the optimal subset since, in the worst case, $\ivFn$ could be adversarial and return a value of $0$ for each of the items selected by the algorithm, and positive utility only for those not selected. Hence, we make some natural assumptions on $\ivFn$ such that the problem becomes tractable. In practice, it is reasonable to assume that $\ivFn$ varies `smoothly' over the candidate set $\groundSet$ such that similar items in $\groundSet$ have similar $\ivFn$ values. In this work, we model this by assuming that the similarity $\kernel(\cand,\cand')$ of any pair of items $\cand,\cand'\in\groundSet$ is given by a positive definite kernel function \citep{Scholkopf01} $\kernel:\groundSet\times\groundSet\rightarrow \mathbb{R}$, and that $\ivFn$ has low ``complexity" as measured by the norm in the Reproducing Kernel Hilbert Space (RKHS) associated with kernel $\kernel$.  
The RKHS $\mathcal{H}_{\kernel}(\groundSet)$ is a complete subspace of $L_2(\groundSet)$ of `smooth' functions with an inner product $\langle \cdot , \cdot \rangle_{\kernel}$ s.t $\langle \ivFn,\kernel(\cand,.)\rangle=\ivFn(\cand) $  for all $\ivFn \in \mathcal{H}_{\kernel}(\groundSet)$. By choosing appropriate kernel functions, we can flexibly handle items of different types (vectors, strings, graphs etc.). We use the notation $\mathbf{K}$ to refer to the $n\times n$ kernel (Gram) matrix obtained by evaluating $\kernel(\cand,\cand')$ for all pairs of items.

%

\paragraph{Explore-Exploit Tradeoff}
Given the regularity assumptions about the unknown function $\ivFn$, the task can be intuitively viewed as one of trading off  exploration and exploitation. That is, we can either greedily utilize our current knowledge of $\ivFn$ by picking the next item predicted to be of high value, or we can choose to pick an item that may not have the highest expected value but most reduces the uncertainty about $\ivFn$ across the other items. This challenge is akin to the dilemma faced in multi-arm bandit problems.  An important difference in our setting, motivated by practical considerations, is that we {\em cannot select the same item multiple times}. As a consequence, classical algorithms for multi-armed bandits (such as UCB1 of \citet{Auer02} or \ucb of \citet{Srinivas12}) cannot be applied, since they require that repeated experimentation with the same ``arm'' is possible. Furthermore, classical bandit algorithms do not allow arms to have different costs.
In fact, our setting is {\em strictly more general} than the bandit setting: We can allow repeated selection of a single item $\cand$ by just creating multiple, identical copies $\cand^{(1)},\cand^{(2)},\dots$ with identical utility (i.e., $\ivFn(\cand^{(1)})=\ivFn(\cand^{(2)})=\dots$), which can be modeled using a suitably chosen kernel.


Nevertheless, we build on ideas from modern bandit algorithms that exploit smoothness assumptions on the payoff function. In particular, \citet{Srinivas12} show how the explore-exploit dilemma can be addressed in settings where, as in our case, the reward function has bounded RKHS norm for a given kernel function $\kernel$.
We interpret the unknown value function $\ivFn$ as a sample from a Gaussian Process (GP) prior \citep{Rasmussen05}, with prior mean 0 and covariance function $\kernel$. 
Consequently, we model the function as a collection of normally distributed random variables, one for each item. They are jointly distributed, such that their covariances are given by the kernel:
$$\Cov\bigl(\ivFn(\cand),\ivFn(\cand')\bigr)=\kernel\bigl(\cand,\cand'\bigr).$$
\looseness -1 This joint distribution then allows us to make predictions about unobserved items via Bayesian inference in the GP model.
Suppose we have already observed feedback $\vobs_t = \{y_1,\dots, y_t\}$  for $t$ items $\subSet_t = \{\cand_1,\dots, \cand_t\}$,  i.e., $y_i=f(\cand_i)+\noise_i$, where $\noise_i$ is independent, zero-mean Gaussian noise with variance $\hat{\sigma}^2$.
Then, for each remaining item $\cand$, its predictive distribution for $\ivFn(\cand)$ is Gaussian, with mean and variance (using noise variance $\hat{\sigma}$, according to our assumptions) given by:
\begin{align}
\mu_t(\cand) &= \mathbf{k}_t(\cand)^T(\mathbf{K}_t + \hat{\sigma}^2 \mathbb{I} )^{-1}\vobs_t \text{,}\label{eq:predmean}\\
\sigma^2_t(\cand) &= \kernel(\cand,\cand) - \mathbf{k}_t(\cand)^T (\mathbf{K}_t + \hat{\sigma}^2 \mathbb{I})\mathbf{k}_t(\cand) \text{,}\label{eq:predvar}
\end{align}
\looseness -1 where $\mathbf{k}_t(\cand) = [\kernel(\cand_1,\cand),\dots,\kernel(\cand_t,\cand)]^T$, $\mathbf{K}_t$ is the positive semi-definite kernel matrix such that for $i,j\leq t$, $\mathbf{K}_{t,i,j}=[\kernel(\cand_i,\cand_j)]$ and $\mathbb{I}$ is the $t \times t$ identity matrix. In Section~\ref{sec:theory}, we show how we can use these predictive distributions to navigate the exploration--exploitation tradeoff. Note that while we propose a Bayesian algorithm (using a GP prior, and Gaussian likelihood), we prove agnostic results about arbitrary functions $f$ with bounded norm, and arbitrary noise bounded by $\hat{\sigma}$.
\newpage
\section{THE UNIFORM COST CASE}
\label{sec:theory}
\begin{algorithm}[tb]
\caption{\algo}
   \label{alg:main}
\begin{algorithmic}
   \STATE {\textbf{Input:}}  Ground Set $\groundSet$, kernel $\kernel$ and budget $B$
   \STATE Initialize selection set $\subSet$
   \FOR{$t = 1, 2, \dots, B$}
   \STATE \textbf{Model Update:} \\\hspace{.5cm}$[\mu_{t-1}(\cdot),\sigma_{t-1}^2(\cdot)] \leftarrow$ GP-Inference$(\kernel,(\subSet,y_{\{1:t-1\}}))$
   \STATE \textbf{Item Selection:} \\\hspace{.5cm}Set $\cand_t \leftarrow \displaystyle \argmax_ {\cand \in \groundSet/  \{ \cand_{1:t-1} \}} \mu_{t-1}(\cand)+\beta_{t}^{1/2}\sigma_{t-1}(\cand)$
   \STATE $\subSet \leftarrow \subSet \cup \{\cand_t\}$
   \STATE Receive feedback $y_t = \ivFn( \cand_t) + \epsilon_t$
   \ENDFOR
\end{algorithmic}
\end{algorithm}
We first provide the solution for the simple case of uniform costs. In this setting, if the values are known, a greedy algorithm adding items of maximal value solves Problem~\eqnref{eqn:knapsack} optimally. Our key idea in the unknown value case is to mimic this greedy algorithm. Instead of greedily adding the item $v$ with highest predicted gain $\mu_{t-1}(v)$, we trade exploration and exploitation by greedily optimizing an optimistic estimate of the item's value.
Concretely, our algorithm \algo for the uniform cost case performs both a model update and selects the next item upon receiving feedback for the current selected item. The model update is performed according to Equations \eqnref{eq:predmean} and \eqnref{eq:predvar}.

For our selection rule, we borrow a key concept from multi-armed bandits: upper confidence bound sampling. Concretely, we choose
\begin{equation}
     \cand_t = \displaystyle \argmax_ {\cand \in \groundSet \setminus  \{ \cand_{1:t-1} \}} \mu_{t-1}(\cand)+\beta_{t}^{1/2}\sigma_{t-1}(\cand), \label{eqn:GPUCBdecisionRule}
\end{equation}
The tradeoff between exploration and exploitation  is implicitly handled by the time varying parameter $\beta_t$ (defined in Theorem \ref{theorem:mainTheorem}) that alters the weighting of the posterior mean (favoring exploitation by selecting items with high expected value) and standard deviation (favoring exploration by selecting items that we are uncertain about). $\beta_t$ is chosen such that $\mu_{t-1}(\cand)+\beta_{t}^{1/2}\sigma_{t-1}(\cand)$ is a high-probability upper bound on $f(\cand)$, explained further below. 


\paragraph{Regret bounds}

We now present bounds on the regret $R_B$ incurred by $\algo$. Crucially, they {\em do not} depend on 
the size of the ground set $|\groundSet|$, but only on a quantity $C_\textbf{K}$ that depends on the task specific kernel capturing the regularity of the utility function over the set of items. Specifically, for a kernel matrix $\textbf{K}$, the quantity $C_\textbf{K}$ is given by:
\begin{equation}
\label{informationContent}
C_\textbf{K} =  \frac{1}{2} \text{log} \mid \mathbb{I} + \hat{\sigma}^{-2} \textbf{K}\mid.
\end{equation}
We now present the main result about \algo in the uniform cost case.
\begin{theorem}
\label{theorem:mainTheorem}
Let $\delta\in (0,1)$. Suppose that the function $\ivFn$ lies in the the RKHS $\mathcal{H}_{\kernel}(\groundSet)$ corresponding to the kernel $\kappa(\cand, \cand')$ with an upper bound on the norm of $\ivFn$ w.r.t.~$\kappa$ given by $R$ (i.e., $||\ivFn||^2_{\kappa} \leq R$). Further suppose that the noise has zero mean conditioned on the history and is bounded by $\hat{\sigma}$ almost surely. Let $\beta_t = 2R + 300 C_\textbf{K} \log^3 (t/\delta)$. Running \algo with a GP prior using mean zero, covariance $\kernel(\cand,\cand')$ and noise model $N(0,\hat{\sigma}^{2})$, we obtain a regret bound of $O^*(\sqrt{B}(R\sqrt{C_\textbf{K}}+C_\textbf{K}))$ w.h.p. Specifically, 
\[
\text{Pr} \{ R_B \leq \sqrt{C_1B  \beta_{B}C_\textbf{K}}  ~~ \forall B \geq 1 \} \geq 1-\delta
\]
where $C_1 = \frac{8}{\text{log}(1+\hat{\sigma}^{-2})}$.

\end{theorem}
The proof of this theorem is presented in the Appendix.
\paragraph{Interpretation of the Theorem}
Theorem~\ref{theorem:mainTheorem}  guarantees that under sufficiently regular $\ivFn$ and suitable choice of $\beta_t$, the average regret compared to the best subset approaches 0 as $B$ increases.  
Our regret bound depends only on the constant $C_\textbf{K}$ rather than the actual size of the set $\groundSet$. 
It is instructive to think of how the value $C_\textbf{K}$ grows as the size of the ground set, $n=|\groundSet|$ increases. As long as the kernel function is bounded, it can be seen that $C_\textbf{K}$ is $O(n)$. For many commonly used kernel functions, however, this quantity grows strictly sublinearly in the number $n$ of elements. For instance, for the popular RBF kernel in $d$ dimensions (that is, $\groundSet \subseteq \mathbb{R}^d$), it holds that $C_\textbf{K} = C_\textbf{K}(n) =  O((\log n)^{d+1})$. Refer \citet{Srinivas12} for this and other analytical bounds for other kernels. 
In any case,  a problem specific $C_\textbf{K}$ can always be computed efficiently using the formula in Equation \eqnref{informationContent}. Further note that as long as we use a universal kernel $\kappa$ (like the commonly used Gaussian kernel), for finite item sets (as we consider here) the RKHS norm $||\ivFn||_{\kappa}$ is always bounded. Hence, Theorem~\ref{theorem:mainTheorem} guarantees that our regret will always be bounded for such kernels, provided we choose a large enough value for $R$.

An important point to be made here is that the value of $\beta_t$ as prescribed by Theorem~\ref{theorem:mainTheorem} is 
chosen very conservatively for sake of the theoretical analysis. For most practical applications, $\beta_t$ can be scaled down to achieve faster convergence and lower regret.

\section{SELECTING DIVERSE SUBSETS}
\label{sec:diversity}
In some cases, we not only seek high cumulative value of the solution set, but also prefer {\em diversity}. This can be the case because we desire  robustness, fairness etc. Formally, we can encode this diversity requirement into the objective function as done in \eqnref{eqn:DivObjective}.
Hereby $\ivFn$ is an {\em unknown} function that operates on individual elements,
while  $\smFn$ is a {\em known} set function that captures the diversity of a subset.  It is natural to model diversity as a submodular function. 
Formally, a set function $\smFn : 2^{\groundSet} \rightarrow \mathbb{R}$ is {\em submodular} if for every $A \subseteq B \subseteq \groundSet $ and $\cand \in \groundSet \setminus B$, it holds that
\begin{equation}
\label{def:submodularity}
\Delta_\smFn(\cand\mid A)\geq \Delta_\smFn(\cand\mid B),
\end{equation}
where $\Delta_\smFn(\cand\mid A)\equiv \smFn(A\cup\{\cand\})-\smFn(A)$ is called the {\em marginal gain} of adding $\cand$ to set $A$. 
$\smFn$ is called {\em monotone}, if, whenever $A\subseteq B$ it holds that $\smFn(A)\leq \smFn(B)$. 

The rationale behind using submodular functions to model diversity is based on the intuition that adding a new element provides less benefit (marginal gain) as the set of similar items already selected increases. Many functions can be chosen to formalize this intuition.  In our setting, a natural monotone submodular objective that captures the similarity as expressed via our kernel, is
\begin{equation}
\smFn(\subSet) = \frac{1}{2} \text{log} \left \vert  (\mathbb{I}+\noisevar^{-2} \Kernel_{\subSet , \subSet}) \right\vert,
\end{equation}
where $\noisevar\geq 0$. We use this objective in our experiments. 
For this choice, the marginal gain of adding an element $\cand$ to a set $\subSet$ is given by:
\begin{equation}
\Delta_D (\cand \mid \subSet) = \frac{1}{2} \text{log} (1+\noisevar^{-2}\sigma_{\cand \mid \subSet}^2),
\end{equation}
where $\sigma_{\cand \mid \subSet}^2$ is the predictive variance of $f(\cand)$ in a GP model, where the values of elements in $\subSet$ have already been observed up to Gaussian noise with variance $\sigma_n^2$. Conveniently, while executing \algo, if $\hat{\sigma}=\noisevar$, we already compute $\sigma_{\cand \mid \subSet}^2$ in order to evaluate the decision rule~\eqnref{eqn:GPUCBdecisionRule}. Hence, at almost no additional cost we can compute the marginal gain in diversity for any candidate item $\cand$.

In order to select items that provide value and diversity, it is natural to modify the selection rule of \algo in the following way:
\begin{align}
     \cand_t = \displaystyle \argmax_ {\cand \in \groundSet \setminus  \{ \cand_{1:t-1} \}} &(1-\lambda)\Bigl[\mu_{t-1}(\cand)+\beta_{t}^{1/2}\sigma_{t-1}(\cand)\Bigr]\nonumber \\+&\lambda\; \Delta_D(\cand\mid\{\cand_1,\dots,\cand_{t-1}\}). \label{eqn:DivDecisionRule}
\end{align}
\looseness -1 This decision rule greedily selects item $\cand$ that maximizes a high-probability upper bound on the marginal gain $\Delta_F(\cand\mid\{\cand_1,\dots,\cand_{t-1}\})$ of the {\em unknown} combined objective $F$.

\paragraph{Regret bound}
The regret bound in Section~\ref{sec:theory} depended on the fact that we were optimizing against $\ivFn$ that assigned values to individual elements, $\cand \in \groundSet$.
The same bounds need not hold in the more challenging setting when trading value against diversity.
In fact, even if both $\ivFn$ and $\smFn$ are completely {\em known} for all $\cand \in \groundSet$, it turns out that optimizing $F$ in \eqnref{eqn:DivObjective} is NP-hard for many monotone submodular functions $\smFn$ \citep{Feige98}. While finding the {\em optimal} set is hard, \citet{Nemhauser78} states that -- for a {\em known} monotone submodular function -- a simple greedy algorithm provides a near-optimal solution.

\begin{figure*}[t!]
\centering 
\subfigure[\emph{Balancing utility and diversity}]{
\includegraphics[width=0.28\textwidth]{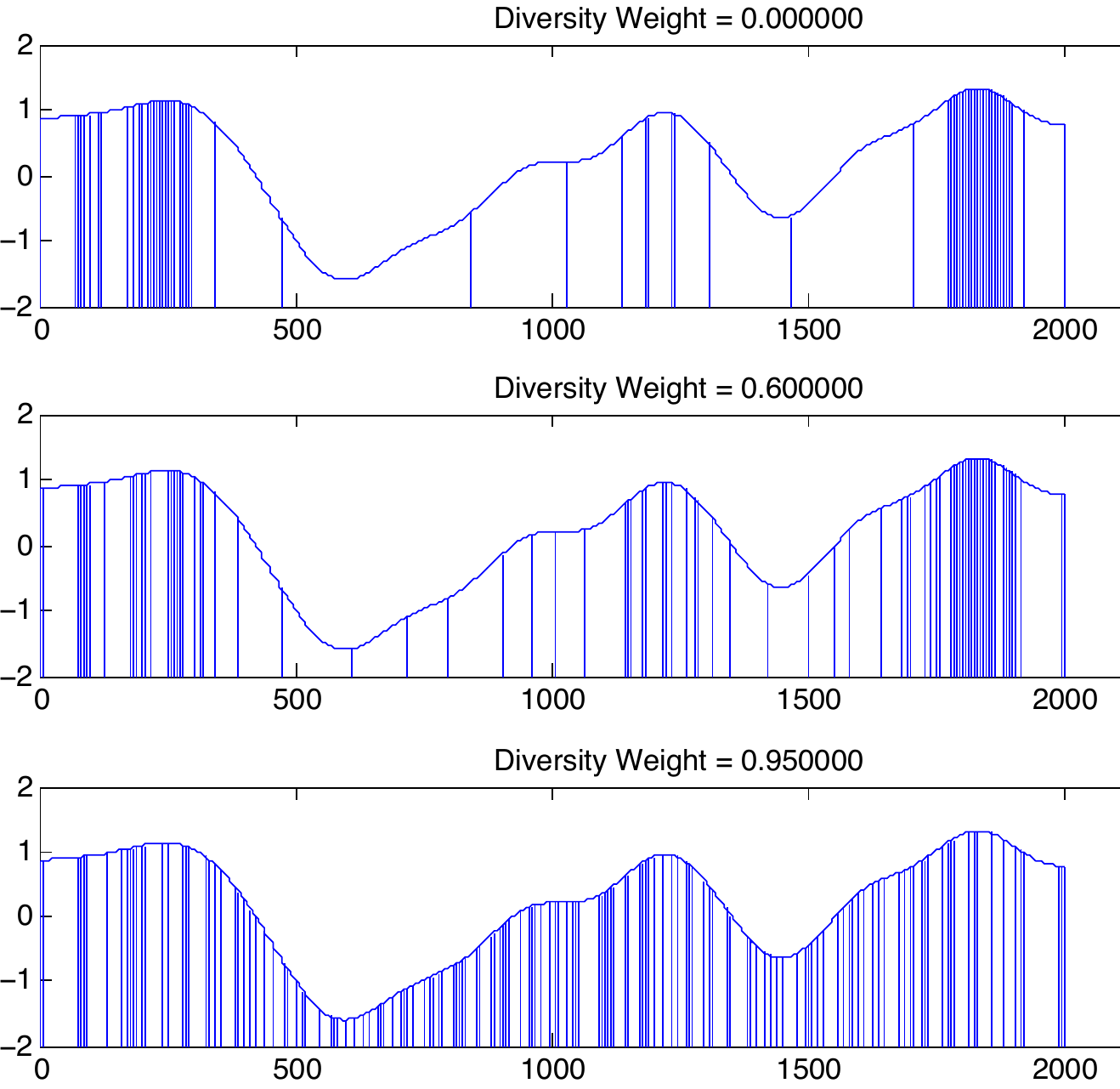}
\label{fig:divIllustration}
 }
 \subfigure[\emph{Average Regret (Diversity)}]{
\includegraphics[ width=0.33\textwidth, height = 0.23\textwidth]{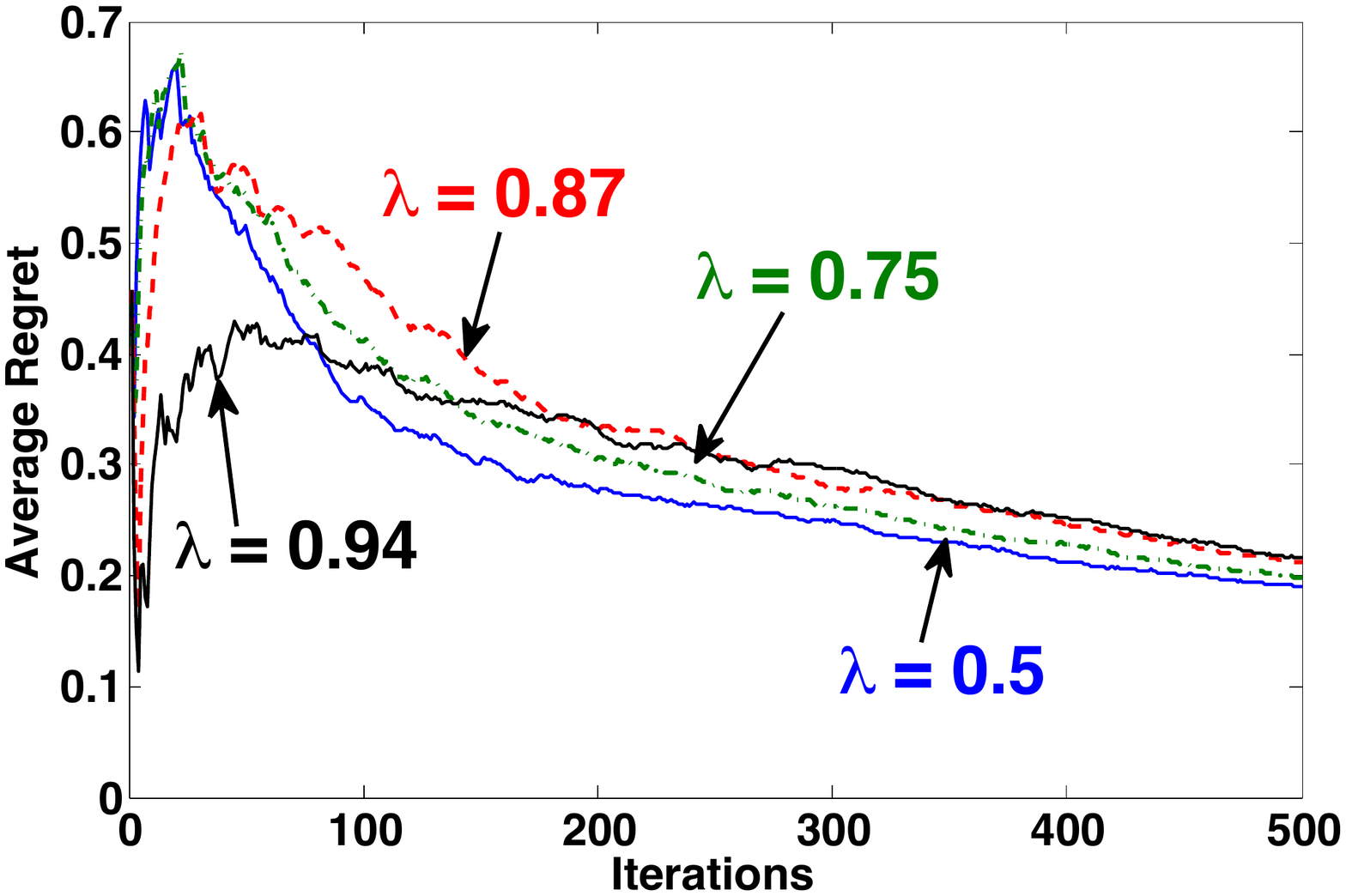}
\label{fig:cumulVal}
 }
\subfigure[\emph{Effect of Inducing Diversity}]{
\includegraphics[width=0.33\textwidth, height = 0.23\textwidth]{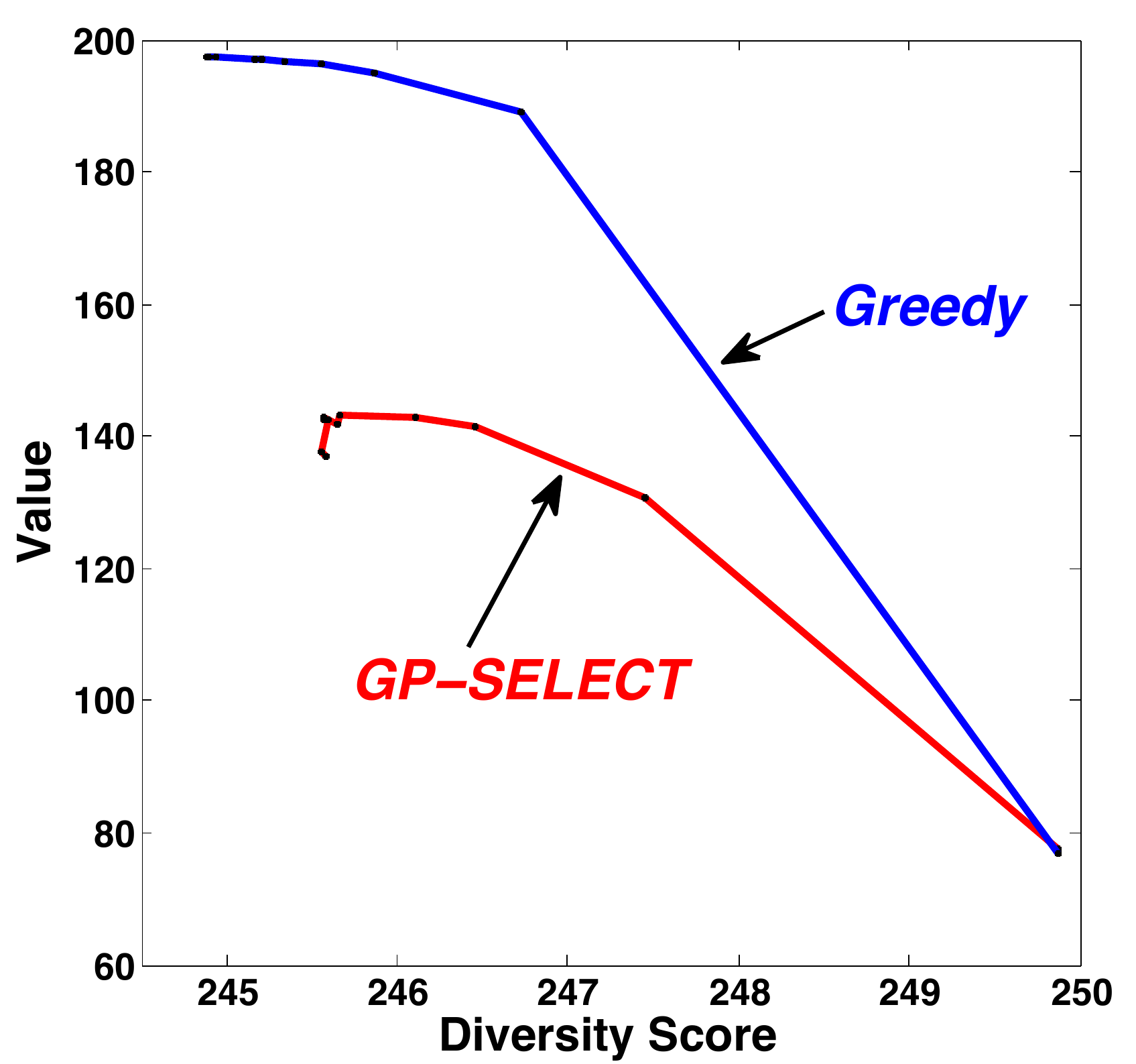}
\label{fig:divTradeoff}
 }
 \\[-3mm]

\caption{
\textbf{\subref{fig:divIllustration}:} Illustration of sets selected for trading $\ivFn$ against $\smFn$ when varying parameter $\lambda$.
\textbf{\subref{fig:cumulVal}:} Performance of \algo in selecting diverse subsets. For different values of $\lambda$, the average regret against the greedy approximate algorithm decreases. 
\textbf{\subref{fig:divTradeoff}:} Improvements in diversity can be obtained at little loss of utility.
}
\label{fig:diversity}
\end{figure*}

%
%

Formally, suppose $S'_0=\emptyset$ and $S'_{i+1}$, the greedy extension to $S'_i$. That is, $S'_{i+1}=S'_i\cup\{\argmax_{\cand\in\groundSet\setminus S'_i}\Delta_F(\cand\mid S'_i)\}$. Thus, $S'_B$ is the set we obtain when selecting $B$ items, always greedily maximizing the marginal gain over the items picked so far. Then it holds that $F(S'_B)\geq (1-1/e)\max_{|S|\leq B} F(S)=(1-1/e) F(S_B^*)$. Moreover, without further assumptions about $\smFn(S)$ and $f$, no efficient algorithm will produce better solutions in general. Since we are interested in computationally efficient algorithms, we measure the regret of a solution $S_B$ by comparing $F(S_B)$ to $F(S'_B)$, which is the bound satisfied by the greedy solution. Formally, $R_B = (1-1/e) F(S_B^*)-F(S_B)$.


%
%
%
%
\begin{theorem}
\label{theorem:divTheorem}
Under the same assumptions and conditions of Theorem \ref{theorem:mainTheorem}, 
\[
\text{Pr} \{ R_B \leq \sqrt{C_1B  \beta_{B} C_\textbf{K}}  ~~ \forall B \geq 1 \} \geq 1-\delta,
\]
where $R_B=(1-1/e) F(S_B^*)-F(S_B)$ is the regret with respect to the value guaranteed when optimizing greedily given full knowledge of $f$ and $D$.
%
\end{theorem}

Please refer to the Appendix for the proof of this theorem. It rests on interpreting \algo as implementing an approximate version of the greedy algorithm maximizing $\Delta_F(\cand\mid S_t)$. In fact, Theorem~\ref{theorem:divTheorem} can be generalized to a large number of settings where the greedy algorithm is known to provide near-optimal solutions for constrained submodular maximization. 

As an illustration of the application of this modified \algo to diverse subset selection, refer to Figure \ref{fig:divIllustration}. When $\lambda = 0$, \algo reverts back to Algorithm \ref{alg:main} and hence, picks locations only based on its expected $\ivFn$ value. This is clear from the thick bands of points sampled near the maximum. At $\lambda =0.6$, \algo balances between expected $\ivFn$ values of the points and the marginal gain in diversity of the points picked $\Delta_D (\cand \mid \subSet)$. At $\lambda$ close to 1, \algo picks mostly by marginal gain which will be approximately uniform if the kernel used is isotropic (e.g. Gaussian kernel). 

\section{NON-UNIFORM COSTS}
\label{sec:knapsack}
In the general case where each element $\cand \in \groundSet$ has different costs of selection $c_\cand$, the budget $B$ is the total cost of all items in the selected subset. 
We modify the selection rule in Algorithm~\ref{alg:main} to take the estimated cost-benefit ratio into account. Most of the other steps remain the same except ensuring that we respect the budget, and the formula for computing $\beta_t$. The new selection rule for the setting without diversity is:
\begin{align}
\cand_{t} = \displaystyle \argmax_ {\cand \in \groundSet \setminus  S , c_\cand \leq B-C(S)} \frac{\mu_{t-1}(\cand)+\beta_{t}^{1/2}\sigma_{t-1}(\cand)}{c_\cand}.
\label{eqn:DecisionRuleKnapsack}
\end{align}
Hence, instead of maximizing an optimistic estimate of the item's value, we greedily maximize an optimistic estimate of the benefit-cost ratio. Note that this greedy rule encourages some natural opportunistic exploration:  Initially, it will select items that we are very uncertain about (large $\sigma_{t-1}$), but that also have little cost. Later on, as the utility is more accurately estimated, it will also invest in more expensive items, as long as their expected value ($\mu_{t-1}$) is high.

\begin{figure*}[t!]
\centering 
\subfigure[\emph{Average Regret}]{
\includegraphics[width=0.32\textwidth, height = 0.23\textwidth]{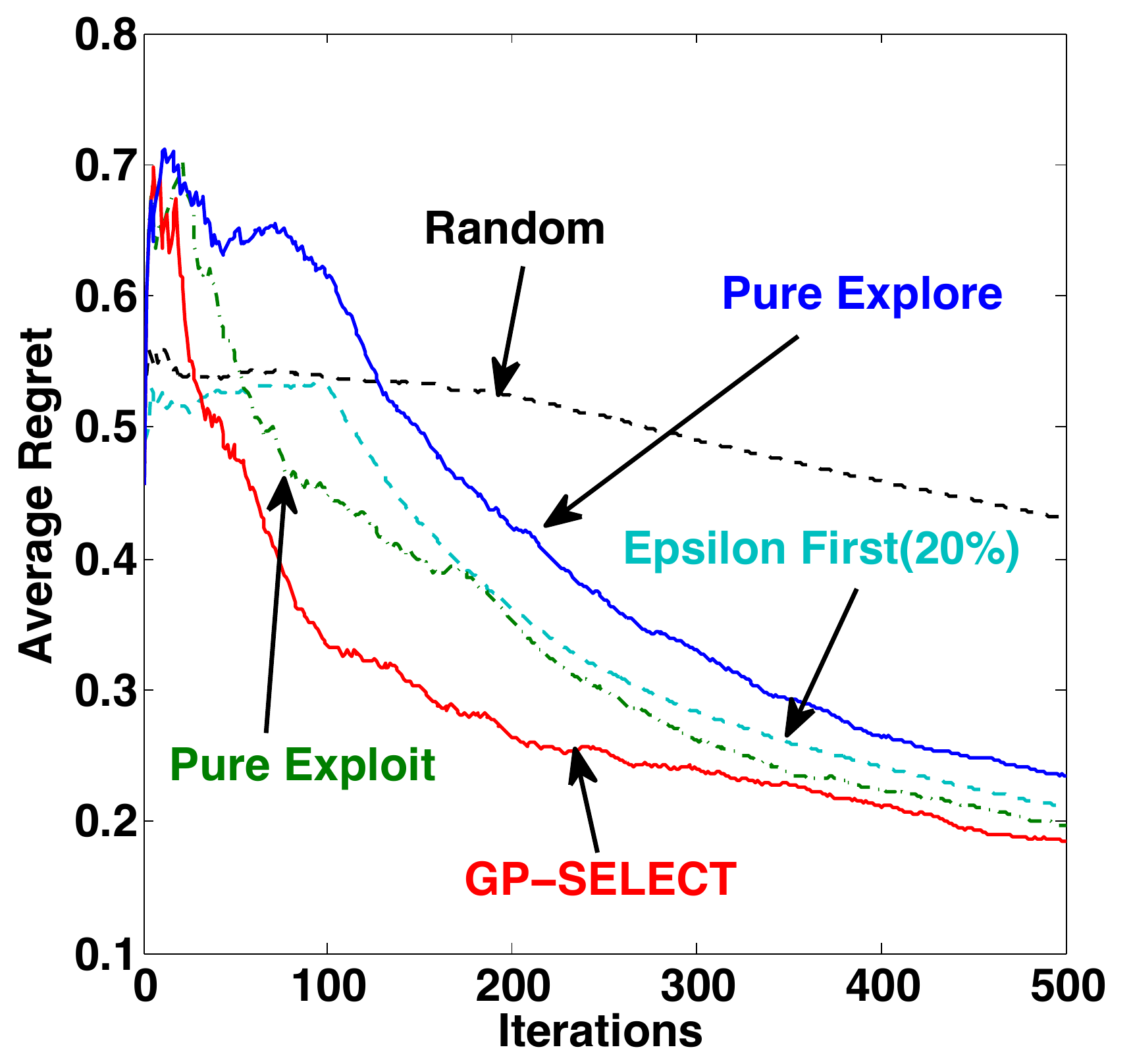}
\label{fig:avgReg}
}
 \subfigure[\emph{Non-uniform costs}]{
\includegraphics[width=0.29\textwidth]{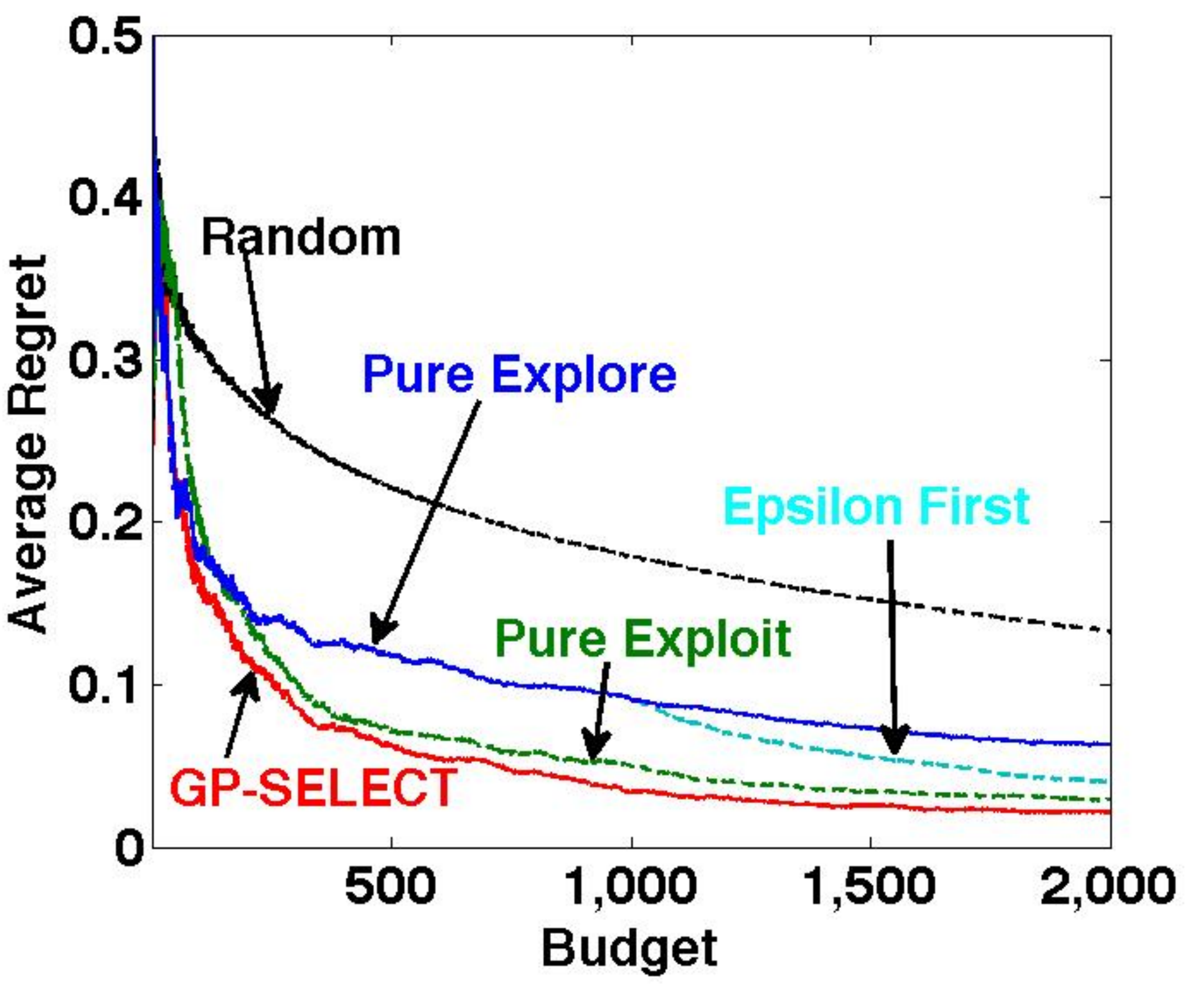}
\label{fig:knapsack}
}
\subfigure[\emph{Flight ticket price change prediction}]{
\includegraphics[width=0.32\textwidth]{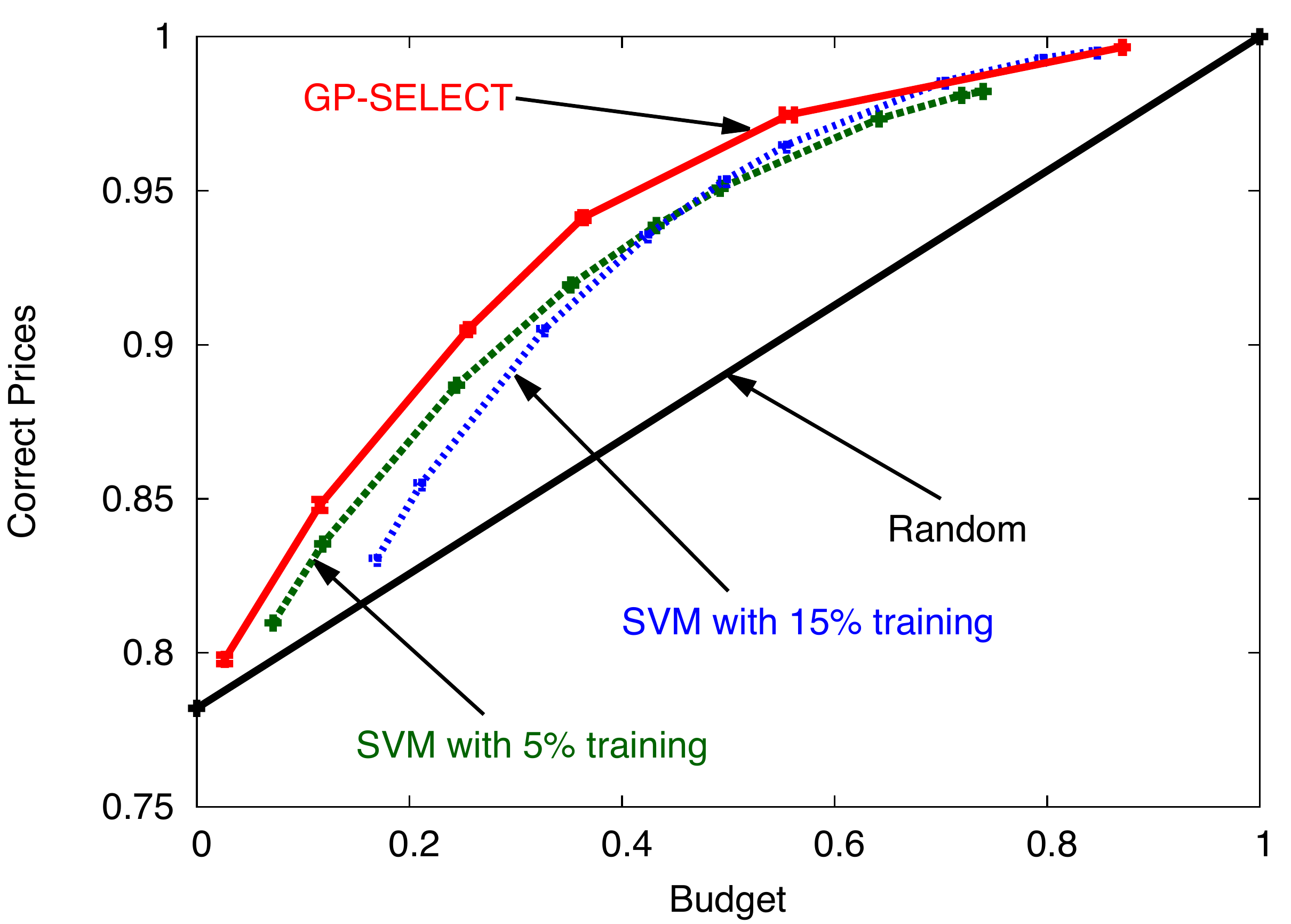}
\label{fig:amadeus}
 } 
 \\[-3mm]
 \caption{
\textbf{\subref{fig:avgReg}:} While average regret decreases for all non-naive algorithms, \algo drops much earlier and continues to outperform the baselines in the vaccine design task.
\textbf{\subref{fig:knapsack}:} Comparison of \algo with the baselines for the vaccine design task under non-uniform item costs.
\textbf{\subref{fig:amadeus}:} \algo outperforms benchmarks on the fare change prediction task.
\vspace{-3mm} }
\label{fig:VacData}
\end{figure*}

The idea above can be generalized to encourage diversity as well.
The selection rule in \eqnref{eqn:DivDecisionRule} can be modified to maximize the ratio
\begin{align}
          \displaystyle \frac{(1-\lambda)\Bigl[\mu_{S-1}(\cand)+\beta_{S}^{1/2}\sigma_{S-1}(\cand)\Bigr] + \lambda \Delta_D(\cand\mid S)}{c_\cand}.
     \label{eqn:DivDecisionRuleKnapsack}
\end{align}
Hence, in this most general setting, we greedily optimize a high-probability upper bound on the cost-benefit ratio of the marginal gain for the joint objective.

Upon these modifications, we can obtain the result presented in Theorem \ref{theorem:divKnapsackTheorem}. The result holds for running \algo for selecting diverse items with items in the ground set having non-uniform costs of selection. Again, we present the proof of the Theorem in the Appendix.

\begin{theorem}
\label{theorem:divKnapsackTheorem}
Under the same assumptions and conditions of Theorem \ref{theorem:mainTheorem}, running \algo with non-uniform costs for the items, we have that
\[
\text{Pr} \{ R_B \leq \left( \displaystyle \max_{\cand \in \groundSet} f(\cand) + c_{max} \sqrt{C_1B  \beta_{B} C_\textbf{K}} \right)  ~~ \forall B \geq 1 \} \geq 1-\delta,
\]
where $R_B=(1-1/e) F(S_B^*)-F(S_B)$ is the regret with respect to the value guaranteed when optimizing greedily given full knowledge of $f$ and $D$.

%
\end{theorem}

\section{EXPERIMENTAL EVALUATION}

\subsection{Case Study I: Airline Price Update Prediction Task}

Amadeus IT group SA\footnote{http://www.amadeus.com} is a Global Distribution System (GDS) for airline prices. One of the services provided by Amadeus is finding the cheapest return fare between cities X and Y on requested dates of travel. This is currently done by frequently querying all the airlines for their respective cheapest fares for each pair of cities and then aggregating the results to maintain this information. This consumes a lot of bandwidth and time. Also, computing the fare for a given request is a computationally expensive task as the cheapest fare might include multiple hops possibly operated by different airlines. Hence, a table of precomputed current best prices is maintained in order to quickly respond to fare requests by customers.  Since the database is typically very large and computing fares is relatively expensive in terms of computation and network bandwidth, it is challenging to frequently recompute all fares (i.e., update the entire table). Since similar prices for similar fare requests (table entries) often change at the same time, the goal is to selectively recompute only entries that changed. This task can be naturally captured in our setting, where items correspond to table entries selected for recomputation, and the utility of an item is 1, if the entry changed and 0, otherwise.


The data provided by Amadeus for this task was collected in December 2011. It consists of cheapest fares computed for 50,000 routes (origin-destination pairs) and for all departure dates up to 90 days into the future. For each departure date, the return date could be up to 15 days after the departure. 
The budget for selection corresponds to the total number of price refresh computations allowed. 
Our performance metric is the ratio between the total number of correct prices (i.e., correct entries in the table) and the total number of prices in the repository. Since we have the data with all the correct prices, we are able to compute the number of prices an algorithm would have missed to update (regret). 

In our experiments, we pool all the data for a given route together, and sequentially process the data set, one ``current date'' at a time. The task is to discover items (table entries) that have changed between the current date and the next date. We thus instantiate one instance of the active discovery problem per route per day. For each instance, we select from 
$90\cdot 15=1350$ prices to recompute.  Typically only 22\% of the data changed between days, hence even with a budget of 0, around 78\% of the prices are correct. In order to capture similarity between items (table entries), we use the following features: {\em date, origin, destination, days until departure, duration of stay, current price}. We use an RBF kernel on these features and tune the bandwidth parameter using data from four routes (origin-destination pairs). We compare \algo against the following baselines:
\begin{enumerate}
\setlength{\itemsep}{0mm}
\item \textbf{Random:} Naive baseline that picks points to query uniformly at random until the budget is exhausted 
\item  \textbf{Epsilon-First:} A Support Vector Machine (SVM) classifier is trained on a randomly sampling part of the data. Concretely, we report the values for two different settings that perform best among other options(5\% and 15\%) of the data. The SVM is then used to predict changes, and the predicted points are updated. When higher budgets are allowed, we use a weighted version of the SVM that penalizes false negatives stronger than false positives.
\end{enumerate}
Figure \ref{fig:VacData} \subref{fig:amadeus} presents the results of our experiments. In general, \algo performs better than the baselines. Note that all three non-naive algorithms reach similar maximum performance as the budget is increased close to 100\% of the total number of items.  

\newpage
\subsection{Case Study II: Vaccine Design Task}
The second task we consider is an experimental design problem in drug design.
The goal is to discover peptide sequences that bind well to major histocompatibility complex molecules (MHC). MHC molecules act as a mediator for interaction of leukocytes (white blood cells) with other leukocytes or body cells and play an important role in the immune system. In our experiments, the goal is to choose peptide sequences for vaccine design that maximizes the binding affinity to these Type I MHC molecules \citep{Peters06peptide}. It is known from past experiments that similar sequences have similar binding affinity responses \citep{Widmer10,Jacob08,Krause11contextual}. 
Instead of selecting only one optimal sequence, it is an important requirement to select multiple sequences as candidates and the actual determination of the best sequence is delayed until more thorough tests are completed further down the drug testing pipeline. Hence, while the task for this dataset can also be viewed as a classification task (binders vs non-binders), we are interested in the actual value of the binding affinity and want to pick a set of peptide sequences that have maximal affinity values. 

The dataset \citep{Peters06peptide}  consists of  peptide sequences of length $l=9$ for the A\_0201 task \citep{Widmer10} which consists of 3089 peptide sequences along with their binding affinities (IC$_{50}$) as well features describing the peptide sequences. We normalize the binding affinities and construct a linear kernel on the peptide features. The task is then to select a subset of up to 500 sequences with maximal affinities. Since this is now inherently a regression task, we used GP regression to estimate the predictive mean of the underlying function. 
The following baseline algorithms were considered for comparison:
\begin{enumerate}
\setlength{\itemsep}{0mm}
\item \textbf{Random:} Naive algorithm that picks sets of size 500 uniformly at random. We repeated this 30 times and report average total affinity values.
\item \textbf{Pure Explore:} This algorithm picks the most uncertain sequence among the remaining sequences. The GP is refitted every time an observation is made. 
\item \textbf{Pure Exploit:} This algorithm always picks the next sequence as the one with the highest expected affinity as computed by GP-regression and the resulting values are used to retrain the GP. This is equivalent to the one-step lookahead policy of \citep{Garnett12}. It is not feasible to implement two or three step lookahead with this large dataset. 
\item \textbf{Epsilon First:} This algorithm randomly explores for a few iterations and then once the GP is trained with the observed responses, behaves exactly like {\it Pure Exploit}. Among all the options we tried, we report results for training on the first 20\% of the budget (100 sequences in this case) since this performed best. A major drawback of this algorithm is that it needs to know the budget a priori. 
We repeated this algorithm 30 times on the data and report the average.
\end{enumerate}
The results of these experiments are presented in Figure \ref{fig:VacData}  \subref{fig:avgReg}, which displays the average regret $R_B/B$. \algo clearly outperforms the baselines in the regret measure. The average regret drops much faster for \algo and continues to remain lower than all the baseline across all the iterations.

\begin{figure*}[t!]
\centering
\subfigure[\emph{Maximizing clicks  on a web-scale recommendation task}]{
\includegraphics[width=0.47\textwidth]{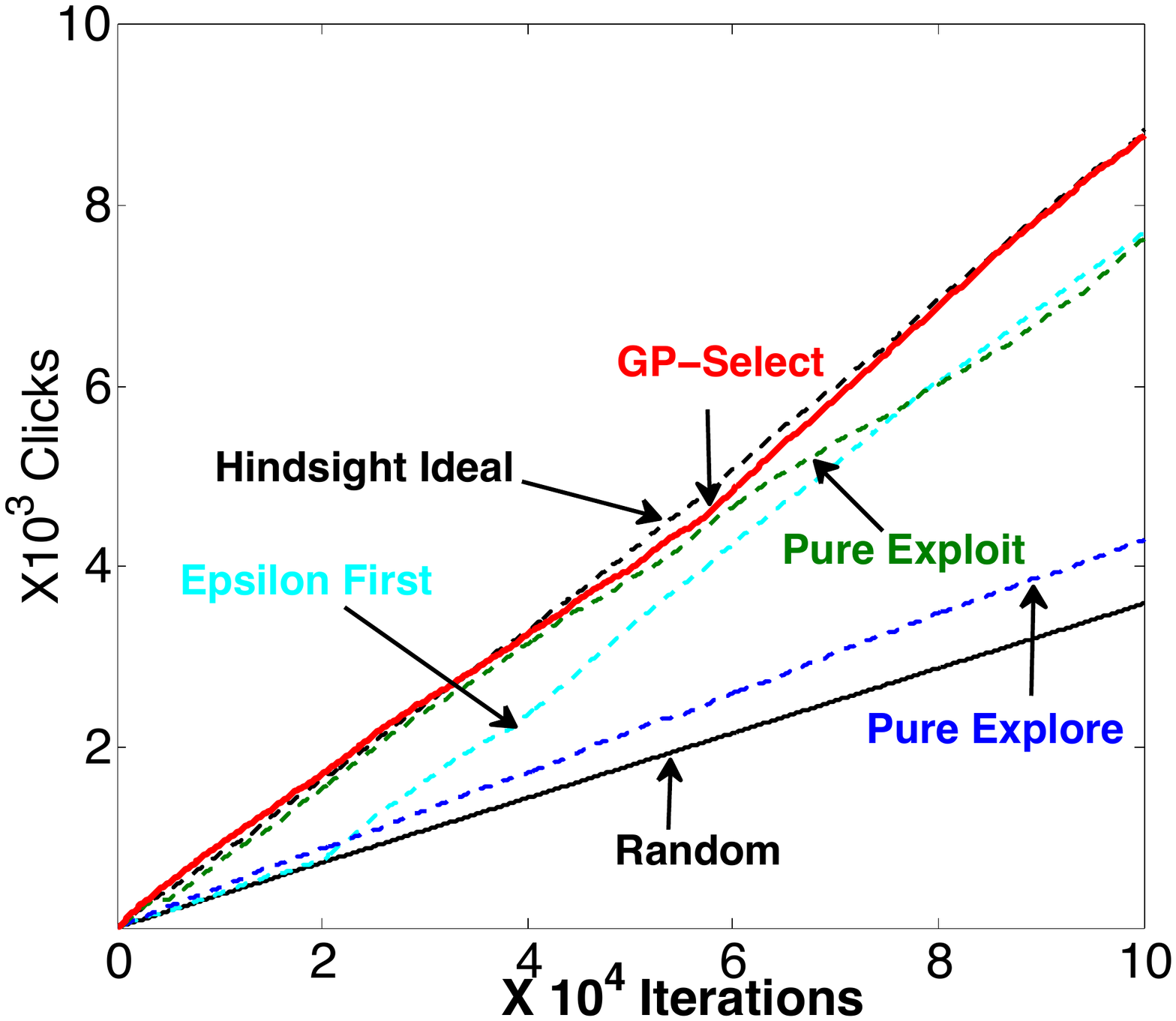}
\label{fig:webscopePlot}
 }
 \hfill
 \subfigure[\emph{Performance Improvements}]{
\begin{tikzpicture}
\clip node (m) [matrix,matrix of nodes,
fill=white,inner sep=0pt,
nodes in empty cells,
nodes={minimum height=1.3cm,minimum width=2.2cm,anchor=center,outer sep=0,font=\sffamily},
column 2/.style={text width=3cm,align=center, row 1/.style={nodes={fill=black!20}}},
column 3/.style={text width=3cm,align=center, row 1/.style={nodes={fill=black!20}}},
column 1/.style={text width=2cm,align=center, nodes={fill=black!20}},
row 1 column 1/.style={nodes={fill=white}},
row 5 column 1/.style={nodes={fill=white}},
row 5/.style={nodes={minimum height=0.5cm,minimum width=2cm,anchor=center,outer sep=0,font=\sffamily}}, 
] {
\pgfmatrixnextcell \textbf{Naive variance update} \pgfmatrixnextcell \textbf{Lazy variance update} \\
 \textbf{Avg. time for one update} \pgfmatrixnextcell 5400ms (for 4m updates) \pgfmatrixnextcell 4.6ms (1 update) \\
 \textbf{Number of updates}  \pgfmatrixnextcell 400 Billion (Predicted) \pgfmatrixnextcell $\sim$ 6 Billion (Actual)\\
 \textbf{Execution Time} \pgfmatrixnextcell 150 hours (Predicted) \pgfmatrixnextcell 3.9 hours (Actual)\\
 \pgfmatrixnextcell   \pgfmatrixnextcell  \\
 };
 \draw (m-4-1.south west) rectangle (m-2-2.north east);
\draw (m-4-2.south west) rectangle (m-1-3.north east);
\draw (m-4-3.south west) rectangle (m-1-3.north east);
\draw(m-1-1.south west) -- (m-1-3.south east);
\draw(m-2-1.south west) -- (m-2-3.south east);
\draw(m-3-1.south west) -- (m-3-3.south east);
\end{tikzpicture}
 \label{table:perfTable}
 } 
 \caption{
Experiments on the news recommendation dataset. \textbf{\subref{fig:webscopePlot}:} \algo outperforms all the baselines by at least 10\% while almost discovering as many clicks (8768) as the hindsight ideal (8863). \textbf{\subref{table:perfTable}: Our failsafe approach for lazy variance updates achieves almost 40X speedup.} 
}
\label{fig:webscope}
\end{figure*}

\paragraph{Choosing Valuable and Diverse Subsets}
Using the same vaccine design dataset, we implement the modified version of \algo presented in Section~\ref{sec:diversity} to select a diverse set of peptide sequences. This requirement of diversity is quite natural for our drug testing application: Very similar sequences, while having similar affinity values, might also suffer from similar shortcomings in later stages of drug testing. We run \algo with different values of the tradeoff parameter $\lambda$, and report the results. Figure \ref{fig:diversity} \subref{fig:cumulVal}, is the average regret $R_B/B$ of \algo for different values of $\lambda$ . The plot demonstrates that when selecting diverse subsets \algo has a similar regret performance as in the initial case when it was selecting only for value. Also, the average regret compared to the greedy optimal solution slightly increases with increase in the value of $\lambda$.  
 Figure \ref{fig:diversity} \subref{fig:divTradeoff} shows the inherent tradeoff between value and diversity. We use values of $\lambda = \{0, 0.5,0.75,0.875,0.9375,0.96875\}$ and plot the performance. We use the diversity function defined in Equation~\eqnref{eqn:DivObjective}. It should be noted that this function is in log scale. From the plot it is clear that for a significant increase in the diversity score, we lose very little functional value, which suggests that robustness of the solution set can be achieved at very little cost.  The {\em greedy} curve on this same plot shows the tradeoff that the greedy algorithm obtains {\em knowing} the utility function. This result serves as a reference, as no efficient algorithm can match it without actually knowing the response function over all the sequences.  Note that as we put all weight on diversity, as expected, \algo's performance converges to that of the greedy algorithm.

\paragraph{Non-Uniform Costs}
The vaccine design task also provides a natural motivation for the non-uniform costs setting. Typically, the cost of testing depends on the actual sequence being tested. Also, field tests differ markedly in their cost of execution. For our dataset, we did not have the costs associated with testing. However, we simulated non-uniform costs for selection of the peptide sequences by sampling $c_v$ uniformly from the range $[c_{min}, c_{max}]$. For different values of $[c_{min}, c_{max}]$, we found that \algo performed better than all the baselines considered. Note that we have used the greedy solution as the hindsight optimal one and this is known to be at most a factor of 2 away from the true optimal solution. While the performance was similar for different values of $c_{min}$ and $c_{max}$, we report results of one of the settings in Figure~\ref{fig:VacData} \subref{fig:knapsack} where $c_{min} = 2$ and $c_{max} = 7$.

\subsection{Case Study III: News Recommendation}
\looseness -1 The Yahoo!~Webscope dataset R6A  \footnote{\url{http://webscope.sandbox.yahoo.com/}} consists of more than 45 million user visits to the \emph{Yahoo! Today} module collected over 10 days in May 2009. The log describes the interaction (view/click) of each user with one randomly chosen article out of 271 articles. It was originally used as an unbiased evaluation benchmark for bandit algorithms \citep{Li10, Vanchinathan14}. Each user $u$ and each article $a$ is described by a 6 dimensional feature vector. That is, $u \in \mathbb{R}^6$ and $a \in \mathbb{R}^6$. Thus, each possible interaction can be represented by a 36 dimensional feature vector (obtained from the vectorized outer product of user and item features) with a click (1) or no-click (0) as the outcome. \citet{Chu09} present a detailed description of the dataset, features and the collection methodology.

In our experiments, we consider an application where we seek to select a subset of articles to present to a subset of users. Hence, we sequentially pick user-item pairs aiming to maximize the number of clicks under a constraint on the number of interactions. Here, a very natural constraint is that we do not want to repeatedly show the same item to the same user.
We randomly subsample 4 million user visits from the Webscope log and treat each interaction as an item with a latent reward that can be observed only when that item is picked. 
As baseline, we also compute the best fixed predictor of the reward given the entire log a~priori. This serves as an unrealistic benchmark to compare our algorithm and other baselines against. We also compare against the other baselines used in the vaccine design task.

For \algo, we use the linear kernel to model similarities between the interactions. This is just the Kronecker product ($\otimes$) of the individual linear kernels on the users and items. We simulate the selection of 100,000 interactions. The total number of clicks in the dataset (of size 4 million) is 143,664, resulting in an average clickthrough rate (CTR) of about 0.0359. 

\paragraph{Results}
\looseness -1 Of the 100,000 selected items, the hindsight-ideal algorithm discovers 8836 items that were eventually clicked on. In comparison, \algo discovers 8768 items beating the other baselines  by at least 10\%. This corresponds to a CTR of 0.0877 which is considerably higher than the average CTR in our dataset. The next best approach is the Epsilon First approach that randomly selects items for 20\% of its budget and then trains a classifier to predict the reward for the remaining items. Detailed results are presented in Figure~\ref{fig:webscope}~\subref{fig:webscopePlot}.

 \subsection*{Scaling to web scale datasets}
 The major bottleneck in using Gaussian Processes is the computation of the posterior mean and variance. There are several works that attempt to speed up GP-based algorithms \citep{Lawrence02,Wang13}, which we can immediately benefit from. Also, our task can be inherently parallelized by distributing the computation across multiple cores/machines and a central processor collects the top UCB scores and picks the one with the best from all the machines. The reward for the chosen item along with the item itself is communicated to the worker nodes which use the information to update the posterior mean and variances.
 
To obtain further improvements, we adapt the idea of lazy variance updates,  originally proposed by \citet{Desautels14} for the bandit setting, and extend it with a novel failsafe variant. We note that the majority of the computation time is spent on computing the posterior variance update, which requires solving a linear system for each item. The key insight is that, for a given item $\cand$, $\sigma_t^2 (\cand)$ is monotonically decreasing in $t$. We exploit this to recompute $\sigma(t)$ only for those items that could influence the selection in round $t$, via use of a priority queue. That is, in every round, we lazily pick the next item $\cand_t$ based on the variance bound from the previous round and update the UCB score for that item. If $\cand_t$ remains the selected item with the new score, we do not need to recompute the variances for the other items. We repeat this process until we find an item whose position at the head of the priority queue does not change after recomputation of the variance. However, note that if we have to recompute for many items in one round, it might be faster to update the variance for items due to the computational overhead associated with using a priority queue (and the benefits of parallelism). Thus, we include a failsafe condition whereby on crossing a machine and task dependent threshold on the number of lazy updates in one round, we switch to the full update. Thus, we eliminate the possibility that a large number of non-contiguous updates might be much slower than one full contiguous update for all the items. Using this technique, we achieve a reduction factor of almost 70 in the number of updates and an overall speedup of 
almost 40 in terms of computational time. The results are presented in Figure \ref{fig:webscope} \subref{table:perfTable}.

\newpage
\section{RELATED WORK}
\label{sec:related} 
\textbf{Frequent itemset mining} \citep{Han07} is an important area of research in  data mining. It attempts to produce subsets of items that occur together often in transactions on a database. However, it is very different in nature from \acron, the problem we address in this paper, since we do not optimize frequency, but (unknown) value.

\textbf{Active learning} algorithms select limited training data in order to train a classifier or regressor. Uncertainty sampling, expected model improvement, expected error reduction, variance reduction are some of the popular metrics in use in this field \citep{Settles12}. 
In (budgeted) active learning, the objective is to learn a function (regression or classification) as well as possible given a limited number of queries. In contrast, we do not seek to learn the function accurately, but only to choose items that maximize the cumulative value (e.g., the number of positive examples) of a function. 
 
\textbf{Active Search} \looseness -1 aims to discover as many members of a given class as possible \citep{Garnett12}. Here, the authors propose single and (computationally expensive) multi-step look ahead policies. It is not clear however how their approach can be applied to regression settings, and how to select diverse sets of items. Furthermore, they do not provide any performance guarantees. \citet{Wang13a} extended this approach to present a myopic greedy algorithm that scales to thousands of items. \citet{Warmuth03} proposed a similar approach based on batch-mode active learning for drug discovery. The algorithms proposed in these works are similar to our exploit-only baseline and further, work only for classification tasks.

\textbf{Multi-arm bandit (MAB) problems}  are sequential decision tasks, where one repeatedly selects among a set of items (``arms''), and obtains noisy estimates of their values \citep{LaiRobbins}. They abstract the explore -- exploit dilemma. In contrast to our setting, in MAB, arms can be selected repeatedly: Choices made do {\em not} restrict arms available in the future. In fact, our setting is a strict generalization of the bandit problem. 
Early approaches like \citet{Auer02} addressed the setting where utilities are considered independent across arms, and hence cannot generalize observations across arms.
More recent approaches \citep{dani,Kleinberg08,Bubeck08} address this shortcoming by exploiting assumptions on the regularity of the utility function. 
In particular, \citet{Srinivas12} develop a bandit algorithm, \gpucb, with regret bounds whenever regularity is captured via a kernel function, which we build on and extend in our work.
In other extensions (e.g. \citep{Kale10,Streeter09online}), the authors consider picking multiple arms per round. However, in these settings, subset selection is a repeated task with the same set of arms available for selection each time. Also, \citet{Kleinberg10Sleeping} consider the case where only a subset of arms are available in each round. However, their results do not apply to our case where arms becomes unavailable upon being selected just once.

\textbf{Stochastic Knapsacks.} Budget limited explore-exploit problems have been studied in context of the stochastic knapsack problem. Hereby, the learning process is constrained by available resources. \citet{Gupta11} provide strong regret bounds for the scalar budget case. 
\citet{Tran-Thanh10} consider prior-free learning for the same problem. \citet{Badanidiyuru13} study the problem under multi-dimensional budget constraints.  However, all approaches consider arms as independent (i.e., uncorrelated), and hence do not generalize observations across similar arms as we do.

\textbf{Submodularity} is a natural notion of diminishing returns of subsequent choices that arises in many applications in machine learning and other domains. A celebrated result about the performance of the greedy algorithm by \citet{Nemhauser78} allows fast yet near-optimal approximation algorithms to a number of  NP-hard problems. However, these approaches assume that the underlying utility function is known, whereas here we attempt to learn it. \citet{Streeter08} use submodular function maximisation to solve online resource allocation tasks.

\textbf{Diversity} inducing rankings and selection have been studied in a variety of settings (e.g. \citep{Slivkins10}). In particular, submodular objective functions are proposed and used by \citet{Streeter09online,Lin11, Yue11, Gunhee11} to model and optimize for diverse solutions. These approaches provide insights on how to quantify preference for diversity via submodularity. However, their algorithms do not apply to our setting, as they consider the setting where sets are repeatedly selected, whereas we build up a single set one element at a time.

\textbf{Lazy variance updates} in explore-exploit settings were proposed by \citet{Desautels14} who generalized the lazy greedy algorithm for submodular maximization \citep{Minoux78}. We have adapted this approach to propose a failsafe lazy variance update technique that gives dramatic speedups in our experiments. 

%

\section{CONCLUSIONS}

We introduced \acronex, a novel problem setting capturing many important real world problems. We presented \algo, a theoretically well founded approach to select high-value subsets from a large pool of items.  We further showed how it can be extended to select diverse subsets, by adding a submodular diversity term to the objective function, and how to handle non-uniform cost.  We prove regret bounds for all these settings.  We further demonstrated the effectiveness on three real world case studies of industrial relevance. To enable the application of \algo to web-scale problems, we proposed a failsafe lazy evaluation technique that dramatically speeds up execution of \algo. Empirically, we find that \algo allows us to obtain a fine-grained tradeoff between value and diversity of the selected items. We believe our results present an important step towards addressing complex, real-world exploration--exploitation tradeoffs.



\paragraph{Acknowledgments} This research was supported in part by SNSF grant \\ 200020\_159557, ERC StG 307036 and a Microsoft Research Faculty Fellowship. The authors wish to thank Christian Widmer for providing the MHC data.
\bibliographystyle{abbrvnat}
\bibliography{refs}

\newpage
\onecolumn
\section*{Appendix}
\textbf{Proof of Theorem~\ref{theorem:mainTheorem}:}

We now prove Theorem~\ref{theorem:mainTheorem}. Our proof builds on the analysis of \cite{Srinivas12}, who address the multi-armed bandit setting with RKHS payoff functions. A difference in our analysis is the usage of the constant $C_\Kernel$ instead of $\gamma_t$. Acoording to the definition in \cite{Srinivas12}, $\gamma_t$  measures the maximum mutual information $I(\ivFn_{\subSet},\vobs_{\subSet})=\frac{1}{2}\log |\mathbb{I}+\sigma^{-2}\mathbf{K}_{\subSet,\subSet}|$ that can be extracted about $\ivFn$ using $t$ samples $\vobs_{\subSet}$ from $\groundSet$. 

\begin{equation}
\label{gammat}
\gamma_B = \displaystyle \max_{\subSet \subset \groundSet, |\subSet|\leq B} I(\ivFn_{\subSet},\vobs_{\subSet})
\end{equation}

But note that the way we have defined $C_\Kernel$, it is easy to see that it is an upper bound on $\gamma_t$. This is because, we can always define a kernel matrix within only the most informative subset of size $t$ (say $\Kernel '$)  and its corresponding $C_{\Kernel '}$ and this would be exactly be $\gamma_t$. And, $C_{\Kernel} \geq C_{\Kernel '}$. This is because given the constraint of our problem setup, after $t$ rounds, the algorithm necessarily has to have picked $t$ distinct items to evaluate. 

Apart from this, there are two important, interrelated changes from the original setting described in \cite{Srinivas12}:
\begin{enumerate}
\item We must respect the additional constraint that we cannot pick the same item twice.
\item The hindsight optimal choice is not a single action but instead a subset of elements in $\groundSet$.
\end{enumerate}
 
With these two changes, in order to prove the statement of the theorem, we need to prove a different statement of Lemma 5.2 from \cite{Srinivas12}. The remaining part of the proof (Theorem 6, Lemmas 5.3 and 5.4) remain the same as in \cite{Srinivas12}.
For the sake of the proof of Theorem \ref{theorem:mainTheorem}, we replace Lemma 5.2 from \cite{Srinivas12} with the following Lemma \ref{lemma:instRegret} and prove a new statement that captures the main differences between the settings. Theorem \ref{theorem:containment}, Lemmas \ref{lemma:infoGainVar} and \ref{lemma:regretBound} are stated without proof and correspond exactly to Theorems 6, 5.3 and 5.4 of \cite{Srinivas12}

The first theorem establishes high probability bounds on the utility function $f$. These carry over without modification.
\begin{theorem}[Srinivas et al., 2012]
\label{theorem:containment} ~ Let $\delta \in (0,1)$. Assume noise variables $\epsilon_t$ are uniformly bounded by  $\hat{\sigma}$. Define:
\[
\beta_t = 2 ||\ivFn||^2_{\kappa} + 300 C_{\Kernel} \log^3 (t/\delta)
\]
Then, $\forall \cand \in \groundSet ,~ \forall B \geq 1 $
\[
|\ivFn(\cand) - \mu_{t-1}(\cand)| \leq \beta_t^{1/2} \sigma_{t_1}(\cand)  
\]
holds with probability $\geq 1-\delta$.
\end{theorem}

The next lemma bounds the instantaneous regret in terms of the widths of the confidence interval at the selected item.
\begin{lemma}
\label{lemma:instRegret}
Fix $ t \in [1,B] $. If $\forall \cand \in \groundSet:\; |\ivFn (\cand) - \mu_{t-1} (\cand) | \leq \beta_t^{1/2} \sigma_{t-1}(\cand) $, then the instantaneous regret $r_t$ is bounded by $2 \beta_t ^ {1/2} \sigma_{t-1}(\cand_t)$.
\end{lemma}

\textbf{Proof:} At any iteration, $t \leq b$,  by the definitions of $\cand_t$ and $\cand^*_t$, one of the following statements is true.
\begin{enumerate}
\item Our algorithm has already picked $\cand^*_t$ in an earlier iteration. In this case, $\exists t'$ s.t $\ivFn(\cand_{t'}^*) \geq \ivFn(\cand_t)$. This is because the ideal ordering has a non-increasing $\ivFn$ value for its elements. Hence,
\begin{align*}
\mu_{t-1}(\cand_t) + \beta_t^{1/2} \sigma_{t-1} (\cand_t) &\geq \mu_{t-1}(\cand^*_{t'}) + \beta_t^{1/2} \sigma_{t-1}(\cand_{t'}^*) \\
                                                                                          & \geq \ivFn(\cand_{t'}^*) \\
                                                                                          & \geq \ivFn(\cand_t^*)
\end{align*}
\item Our algorithm has not yet picked $\cand^*_t$ in an earlier iteration. In this case, 
\begin{align*}
\mu_{t-1}(\cand_t) + \beta_t^{1/2} \sigma_{t-1} (\cand_t) &\geq \mu_{t-1}(\cand^*_t) + \beta_t^{1/2} \sigma_{t-1}(\cand_t^*) \\
                									     &\geq \ivFn(\cand_t^*)
\end{align*}
\end{enumerate}

Thus, in both cases, the statement of the lemma holds.

\begin{lemma}[Srinivas et al., 2012]
\label{lemma:infoGainVar}
The information gain for the objects selected can be expressed in terms of the predictive variances. If $\mathbf{\ivFn}_B = (\ivFn(\cand_t)) \in \mathbb{R}^B$:
\[
I(y_T;\mathbf{\ivFn}_B) = \frac{1}{2} \displaystyle \sum_{t=1}^{B} log(1+\hat{\sigma}^{-2}\sigma_{t-1}^2(\cand_t))
\]
\end{lemma}

\begin{lemma}[Srinivas et al., 2012]
\label{lemma:regretBound}
Pick $\delta \in (0,1)$ and let $\beta_t$ be as defined in Theorem \ref{theorem:containment}. Then, the following holds with probability $\geq 1-\delta$
\[
\displaystyle \sum_{t=1}^B r_t^2 \leq \beta_b C_1 I(y_b);\textbf{f}_b) \leq C_1 \beta_b C_{\Kernel} \text{    }    \forall b \geq 1
\]
\end{lemma}

Now, using Cauchy-Schwartz inequality, $R_B^2 \leq B \sum_{t=1}^B r_t^2$ and this proves the statement of Theorem \ref{theorem:mainTheorem}

\textbf{Proof of Theorem \ref{theorem:divTheorem}:}

We use the proof techniques of \citep{Nemhauser78} and its extension \footnote{A. Krause, A. Singh, and C. Guestrin. Near-optimal sensor placments in gaussian processes: Theory, efficient algorithms and empirical studies. JMLR, 9:235-284, Feb 2008}.

Denote by $\subSet_i=\{\cand_1,\dots \cand_i\}$ the solution set of \algo after $i$ iterations and  by $\subSet^*_i =\{\cand^*_1,\dots \cand^*_i\}$, the solution set of the exact optimal solution after $i$ iterations. 

Given that $\mFn(\subSet)= (1-\lambda)\displaystyle\sum_{\cand \in \subSet} \ivFn (\cand) + \lambda \smFn(\subSet)$, the marginal gain of \algo in the $(i+1)^{\text{th}}$ step is given by:
\[
\Delta_i = \mFn(\subSet_i \cup \{\cand_{i+1}\}) - \mFn(\subSet_i).
\]

Now, from Lemma \ref{lemma:instRegret} and submodularity, in each iteration, $\Delta_i$ can differ from the best greedy choice by at most the width of the  confidence interval

\[
\Delta_i \geq \displaystyle \max_{\cand \in \groundSet \setminus \{ \cand_1 \dots \cand_i\}} \left\lbrace\mFn(\subSet_{i-1} \cup \{\cand\}) - \mFn(\subSet_{i-1}) - \underbrace{(1-\lambda) w_i(\cand_i)}_{\epsilon_{i-1}}\right\rbrace
\]

where $w_i(\cand_i) = 2\beta_i^{1/2}\sigma_i(\cand_{i})$.

Since $\mFn$ is monotone, 
\[
\mFn(\subSet_i \cup \subSet^*_B ) \geq \mFn( \subSet^*_B)
\]

But also, by definition of $\subSet^*_B$, for all $i=0, \dots, B$,
\[
\mFn(\subSet_i \cup \subSet^*_B ) \leq \mFn( \subSet_i) + B(\Delta_{i+1} + \epsilon_i) = \displaystyle \sum_{j=1}^i \Delta_j  + B(\Delta_{i+1}+\epsilon_i)
\]

We can then get the following inequalities,
\[
\mFn( \subSet^*_B ) \leq B(\Delta_{1} + \epsilon_0) 
\]
\[
\mFn( \subSet^*_B ) \leq \Delta_{1}+B(\Delta_{2} + \epsilon_1) 
\]
\[
\vdots
\]
\[
\mFn( \subSet^*_B ) \leq \sum_{j=1}^{B-1} \Delta_j  + B(\Delta_B+\epsilon_{B-1})
\]

Multiplying both sides of the $i^{\text{th}}$ inequality by $(1-\frac{1}{B})^{B-1}$, and adding all the inequalities, we get

\[
\left(\displaystyle \sum_{i=0}^{B-1} (1-1/B)^i \right) \mFn( \subSet^*_B ) \leq B \displaystyle \sum_{i=1}^{B} \left(\Delta_i + \epsilon_{i-1} \right)= B\left( \mFn(\subSet_B) - \underbrace{\sum_{i=0}^{B-1} \epsilon_i}_{R_B} \right)
\]

Further, we can simplify this to,
\[
\mFn(\subSet_B) - R_B \geq \left(1-(1-1/B)^B \right) \mFn(\subSet^*_B) \geq (1-1/e)\mFn(\subSet^*_B)
\]

From Theorem \ref{theorem:mainTheorem}, we can bound $R_B = \sum_{i=0}^{B-1} \epsilon_i$ by \\ $\sqrt{ C_1 B \beta_B C_{\Kernel} } \text{     }  \forall B \geq 1$, thus proving the claim of Theorem \ref{theorem:divTheorem}.

\textbf{Proof of Theorem \ref{theorem:divKnapsackTheorem}:} 
(For ease of presentation we use $c_j$ to denote $c_{v_j}$ when there is no confusion). Also, without loss of generality, we assume that $c_{min} \geq 1$
Our proof is adapted from \cite{Streeter08}. We consider a modified version of greedy algorithm that is allowed to pick from only those elements whose individual costs are less than the budget $B$. Let $(S_j)_j$ be the sequence of subsets chosen by this greedy algorithm. $S_1 \subset S_2 \subset S_3 \dots$. Let $l$ be the maximum index such that $C(S_l) \leq B$. We will show that $F(S_{l+1})$ is nearly optimal. And then, it is easy to see that $F(S_l) \geq F(S_{l+1}) - \displaystyle \max_{\cand \in \groundSet} f(\cand)$ . In order to prove the theorem, we require the following lemma.  

\begin{lemma}
\label{lemma:marginalGreedy}
If $F$ is submodular, $S^* \in \groundSet$ is the optimal subset under budget $B$, and we run the modified greedy procedure picking elements $\{ v_1, v_2, \dots \}$ in that order resulting in sets $S_1 \subset S_2 \subset S_3 \dots$. Then,
\[
F(S^*) \leq F(S_j) + B s_{j+1} + \frac{B}{c_{j+1}} \epsilon_{j+1}
\]
where $s_{j+1} = \frac{F(S_{j+1} - F(S_j))}{c_{j+1}}$ and $\epsilon_{j+1}$ is the error in estimating $s_j$. 
\end{lemma}
\textbf{Proof:}
Let $S^* \setminus S_j = \{o_1,o_2, \dots o_m\}$

Then, 
\begin{align*}
F(S^*) &\leq F(S_j \cup S^*) \\
         &\leq F(S_j) + \displaystyle \sum_{i=1}^m \Delta(o_i \mid S_j)\\
	&\leq F(S_j) + B \left[ \frac{F(S_{j+1}) - F(S_{j}) + \epsilon_{j+1}}{c_{j+1}} \right]\\
	&= F(S_j) + B s_{j+1} + \frac{B}{c_{j+1}} \epsilon_{j+1}\\
\end{align*}
where the second inequality is due to submodularity and the third inequality is due to the greedy selection rule.

For running \algo, $\epsilon_{j+1}$ is instantaneous regret which is upper bounded by the width of the confidence interval, $2 \beta_{S_j} ^ {1/2} \sigma_{S_{j-1}}(v_j)$

Now we are ready to prove Theorem \ref{theorem:divKnapsackTheorem}

Consider $S_{l+1}$, the result of greedy algorithm that just becomes infeasible. Let $\Delta_j = F(S^*) - F(S_j)$
\begin{align*}
\Delta_j &\leq B s_{j+1} + \frac{B}{c_{j+1}} \epsilon_{j+1} \text{(              From Lemma \ref{lemma:marginalGreedy})}\\
		       &=B \left( \frac{\Delta_j - \Delta_{j+1}}{c_{j+1}} + \epsilon_{j+1}\right)
\end{align*}

Rearranging the terms, we get, 
$\Delta_{j+1} \leq \Delta_j \left( 1- \frac{c_{j+1}}{B}\right) + c_{j+1}\epsilon_{j+1}$

Using the fact that $1-\frac{c_{j+1}}{B} \leq 1$, and using the telescopic sum, we get,

$\Delta_{l+1} \leq \Delta_1 \left( \displaystyle \prod_{j=1}^{l} 1 - \frac{c_{j+1}}{B} \right) + \sum_{j=1}^{l} (c_{j+1} \epsilon_{j+1})$

Note that the product series is maximised when $c_{j+1} = \frac{B}{l}$. Thus,

\begin{align*}
\Delta_{l+1} &\leq \Delta_1 \left( 1-\frac{1}{l}\right)^l + \sum_{j=1}^{l} (c_{j+1} \epsilon_{j+1}) \\
                   &\leq  \Delta_1 \frac{1}{e} + \sum_{j=1}^{l} (c_{j+1} \epsilon_{j+1}) \\
                   &\leq F(S^*) \frac{1}{e} + \sum_{j=1}^{l} (c_{j+1} \epsilon_{j+1}) \\
		  &\leq F(S^*) \frac{1}{e} + c_{max} \sum_{j=1}^{l}  \epsilon_{j+1} \\
		  &\leq F(S^*) \frac{1}{e} + c_{max} R_B \\	
\end{align*}

Thus, $F(S_{l+1}) > (1-\frac{1}{e}) F(S^*) - c_{max} R_B$ and $F(S_l) \geq F(S_{l+1}) - \displaystyle \max_{\cand \in \groundSet} f(\cand)$

\end{document}